\documentclass[11pt,a4paper]{article}
\usepackage[final]{acl}
\usepackage[utf8]{inputenc}
\usepackage{graphicx}
\usepackage{xcolor}
\usepackage{enumerate}
\usepackage{footnote}
\usepackage{listings}
\usepackage{verbatim}
\usepackage{svg}
\usepackage[skip=0pt]{caption}
\usepackage[skip=0pt]{subcaption}
\definecolor{pinegreen}{rgb}{0.0, 0.47, 0.44}

\makeatletter
\newcommand\footnoteref[1]{\protected@xdef\@thefnmark{\ref{#1}}\@footnotemark}
\makeatother
\newcommand{\attr}[1]{{\tt \small #1}}
\newcommand{\todo}[1]{\textcolor{red}{[#1]}}
\newcommand{\qb}[1]{\textcolor{blue}{#1}}
\newcommand{\mv}[1]{\textcolor{orange}{#1}}
\newcommand{\framework}{WEBDial}

\title{\framework, a Multidomain, Multitask Statistical Dialogue Framework with RDF
}

\author{Morgan Veyret, Jean-Baptiste Duchene, Kekeli Afonouvi, \\ \textbf{Quentin Brabant, Gw\'enol\'e Lecorv\'e and Lina M.~Rojas-Barahona}\\
 Orange Innovation, 2 Avenue de Pierre Marzin, Lannion, France \\
  \texttt{\{morgan.veyret, jean-baptiste.duchene, kekeli.afonouvi,}\\ \texttt{quentin.brabant,gwenole.lecorve,linamaria.rojasbarahona\}@orange.com} \\}

\date{}

\begin{document}

\maketitle
\begin{abstract}
Typically available dialogue frameworks have adopted a semantic representation based on dialogue-acts\footnote{\label{dact}Dialogue-acts refer to the actions performed by speakers when uttering sentences (\textit{e.g.} informing, requesting, etc.) ~\cite{austin1975things,jurafskyspeech}} and slot-value pairs.
Despite its simplicity, this representation has disadvantages such as the lack of expressivity, scalability and explainability.
We present \framework: a dialogue framework that relies on a graph formalism by using RDF triples instead of slot-value pairs. 
We describe its overall architecture and the graph-based semantic representation.
We show its applicability from simple to complex applications, by varying the complexity of domains and tasks: from single domain and tasks to multiple domains and complex tasks.

\end{abstract}
\section{Introduction}
In the widely adopted slot-value framework, the intent of a sentence is formally represented by predicates associated with sets of arguments, that are respectively called
the dialogue-act\footnoteref{dact} and the slot-value pairs~\cite{young2007cued}.

Despite its simplicity, this ``flat'' representation has several limitations.
Firstly, in term of expressivity: it is not straightforward to express coreferences or to exploit hierarchical relations between arguments using only slot-value pairs. For example, suppose the user said: ``\textit{I want to book a hotel and a restaurant in the same area but not in the same price range}". Although we do not know yet the area nor the price-range, the user has provided important constraints that cannot be expressed with classical slot-value pairs.
Secondly, in terms of scalability: it is difficult to integrate new domains and to adapt them to this representation. 
Finally, in terms of explainability: this simple representation limits causal relations that can be used to explain decisions.

In order to overcome these limitations, we present \framework: a dialogue framework which uses a semantic representation based on RDF graphs.
These graphs are built upon an ontology where the dialogue, domain and task knowledge can be seen as separated blocks, which should ease the burden of integrating new domains and tasks.
We prototyped four applications within this framework:
(i) single domain, single task; 
(ii) multi-domain, multi-task; 
(iii) WEB domains, single task 
and (iv) single-domain, complex task. 


\section{Related Work}
Unlike pure symbolic frameworks \cite{wessel_ontovpaontology-based_2019, walker_dialogue_2021}, we aim to keep the advantages of statistical dialogue systems~\cite{ultes2017pydial} to easily integrate the recent advances in deep and reinforcement learning.

Our original goal was to integrate our RDF-based approach into an existing dialogue framework. We studied six of them:
RASA~\cite{bocklisch2017rasa}, PyDial~\cite{ultes2017pydial}, ConvLab2~\cite{zhu2020convlab}, ParlAI~\cite{miller2017parlai}, DeepPavlov~\cite{burtsev2018deeppavlov} and Plato~\cite{papangelis2020plato}. We evaluated the possibility of extending these existing framework with RDF and looked at the following features: modularity, quick prototyping, RDF support, model availability and reinforcement learning (RL) support. We also were interested in a research oriented framework in which an active communitiy is involved. Only RASA and ParlAI have an active community, they were partially modular, in most of them it is possible to quickly implement a proof of concept, models are available for many of them but RL is not supported in RASA nor in DeepPavlov.
ConvLAB2 is highly tighten to the dataset while ParlAI is oriented towards end-to-end applications, without supporting a modular architecture.
None of these frameworks support RDF out-of-the-box or allow easy modification to add its support.

After this analysis we decided to design and implement our own proof of concept. This gave us more space to explore and refine our ideas.

\section{Architectural Description}
\label{ss:archi_high}
\begin{figure}[t!]
	\centerline{\includegraphics[width=\linewidth]{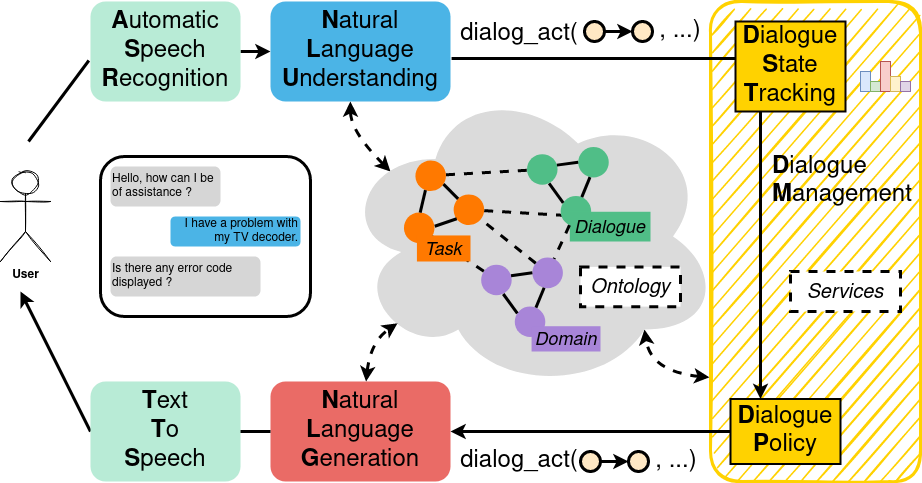}}
	\vspace{2mm}
	\caption{\small\noindent High-level depiction of the proposed dialogue framework. Arrows indicate data flow.} 
	\label{f:architecture}
\end{figure}

\framework~is a modular dialogue system (Figure \ref{f:architecture}), its architecture consists of: a speech-processing front-end, natural language understanding (NLU), a dialogue-state tracker (DST), a dialogue policy (DP), natural language generation (NLG), and a speech synthesis component.

A transcribed utterance is passed to a NLU module which returns a semantic formalism. In turn this semantic formalism passes to  a DST component which returns the probability distribution among the information necessary to properly respond (\textit{e.g.} the belief state). The DP decides of the best system action and a NLG component verbalises it.
Note that, in practice, several successive modules can be implemented in one unique model (\textit{e.g.} DST and DP or DP and NLG).
Similarly, one model may be skipped (\textit{e.g.} using directly DST without calling the NLU) or implemented by multiple sub-components (\textit{e.g.} multi-stage NLU).

\paragraph{NLU} 
The output of the NLU component is expressed in the formalism proposed in Section~\ref{s:sem}. The understanding component can rely on a semantic parser or a neural model. Depending on the application and thanks to \framework~modularity, we were able to reuse existing neural models~\cite{zhu2020convlab} .

\paragraph{DST} The DST module as defined in the literature maintains a dialogue state, which is typically a set of variables denoting the slots that the system must fill-in to complete the task.
Two options are possible: (i) RDF to slot-value mapping: map the RDF representation to slot-value pairs by using rules or human annotations or (ii) dense state representation: learn a representation in a neural network from the RDF inputs. In the implemented applications, we used the first option.

\paragraph{DP}
The DP uses the belief state to select the next action. Its output follows the formalism proposed in Section~\ref{s:sem}. We implemented handcrafted policies for the tasks of our applications.

\paragraph{NLG}
The NLG component translates the action selected by the DP into natural language. This can be achieved in multiple ways, using for example template-based rules or a neural model. For our applications, we implemented a template system to work with our proposed formalism. We also reuse existing models for some applications~\cite{zhu2020convlab}. 
The modularity of the framework allows to easily integrate more powerful generation models such as those presented in \cite{webnlg-2020-international}.

These components rely on concepts defined in an ontology to achieve the dialogue task as well as services providing access to specific functionalities such as querying a database or calling remote APIs.

\section{Ontology}
\label{s:onto}
\begin{figure}[t!]
	\includegraphics[width=\linewidth]{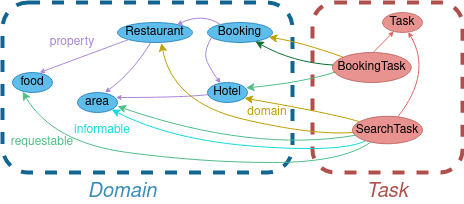}
	\caption{\small\noindent Excerpt of the ontology of Multiwoz.}
	\label{f:onto}
\end{figure}

In slot-value based systems, the scope of possible conversations is defined by the available slots and their allowed values; in \framework~ it is defined by the concepts and the relations in the ontology.
For instance, our MultiWoz ontology (Figure~\ref{f:onto}) has a Hotel concept and properties such as \textit{address} and \textit{price range}. It also has a Booking concept, and a property for relating Booking instances to Hotel instances.

We distinguish three parts in the ontologies: dialogue-related, task-related and domain-related.
The dialogue part defines things shared across tasks and domains such as the concept of User, System or DialogueTurn. The task part exposes information specific to the task that may be reused across domains such as the informable or the requestable properties for an information seeking task.
The domain part defines concepts that are limited to a specific domain such as a Hotel or a Restaurant.
This separation facilitates the reuse of concepts between different applications. For example, information seeking concepts are used in both the restaurant search application and the movie search one.

For each application, all the concepts are linked together as shown in Figure \ref{f:onto} to construct a description of the world manipulated by the system upon which a semantic representation is created.

\section{Semantic Representation}
\label{s:sem}



\begin{figure}[t!]
\centering
\includegraphics[width=\columnwidth]{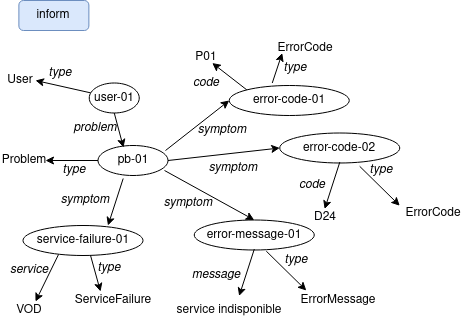}
\caption{Semantics for the user's input: ``\textsl{I no longer have video on-demand. The screen displays unavailable service. It also displays error codes PO1 and D24}''. The dialogue act is \textit{inform}.}
\label{f:nlusem}
\end{figure}

\begin{figure}[t!]
\centering
 \includegraphics[width=\columnwidth]{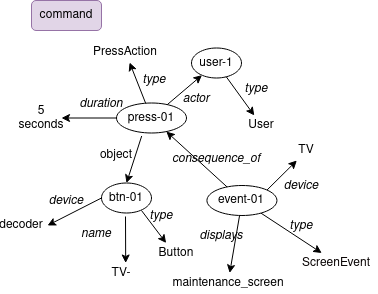}
\caption{Semantics for the system's instruction: ``\textsl{Press the button on the front of the decoder for approximately 5 seconds then the configuration screen of your decoder is displayed}''. The dialogue act is \textit{command}.}
\label{f:nlgsem}
\end{figure}

Our semantic representation of intents and actions defines sequences of dialogue acts with their arguments.
These arguments are graphs where edges and vertices refer to concepts and properties from the ontology.
Figures \ref{f:nlusem} and \ref{f:nlgsem} show examples of using this representation for utterances of an HelpDesk task.

An excerpt of the underlying grammar for this formalism is presented in Figure \ref{f:sem}. It defines domain-independent non-terminals such as \attr{INTENT}~\footnote{The user's intention is represented by the \attr{"INTENT"} in agreement with ~\cite{jurafskyspeech}.},
\attr{DIALOGE\_ACT}\footnoteref{dact} and \attr{TRIPLE}. 
Some non-terminals can be domain dependent such as \attr{ACTION}, \attr{PROPERTY} and \attr{OBJ}.
The grammar embeds RDF triples allowing the use of variables to represent instances (\textit{e.g.} \texttt{error-code-01} and \texttt{error-code-02} in Figure \ref{f:nlusem}).
\begin{figure}[t]
\begin{lstlisting}[basicstyle=\ttfamily\scriptsize,keepspaces=true,breaklines=false]
INTENT-> (DIALOGUE_ACT"("ARGS*")")+
ARGS->  TRIPLE+ | COORD"(" "[" TRIPLE+ "]"+ ")"
COORD-> simultaneous | sequential | alternative 
DIALOGUE_ACT-> inform | request | command | 
        confirm | ...
TRIPLE-> "("ENTITY "," PROPERTY "," VAL|ENTITY ")"
ENTITY-> OBJ|ACTOR|ACTION
ACTOR-> user | system
VAL-> numeric | alphanumeric | ...

!\color{pinegreen}ACTION!-> !\color{pinegreen}press! | !\color{pinegreen}unplug! | !\color{pinegreen}restart! !\color{pinegreen}...!
!\color{pinegreen}PROPERTY!-> !\color{pinegreen}problem!| !\color{pinegreen}symptom! |  !\color{pinegreen}error!| !\color{pinegreen}...!
!\color{pinegreen}OBJ!->!\color{pinegreen}tv-screen! | !\color{pinegreen}decoder! | !\color{pinegreen}hdmi-cable! | !\color{pinegreen}...!
\end{lstlisting}
    
\caption{Excerpt of the grammar for the proposed meaning representation
}
\label{f:sem}
\end{figure}

This formalism seems adapted to bear complex instructions, coreferences and constraints through the use of variables that further define instances. We released RDFDial\footnote{RDFDial is publicly available in: \url{https://github.com/Orange-OpenSource/rdfdial}}, a dataset with dialogues annotated with this formalism. It contains the datasets San Francisco Restaurants and Hotels\footnote{\url{ https://www.repository.cam.ac.uk/items/62011578-23d4-4355-8878-5a150fb72b43}}, DSTC-2\footnote{\url{https://github.com/matthen/dstc}} and MultiWoz2.3\footnote{ https://github.com/thu-coai/ConvLab-2/tree/master/data/multiwoz2.3}  as well as a synthetic data generated with our framework by running dialogues with a simulated user and the dialogue system.

\section{Applications}
Four applications were implemented within \framework: 
\begin{enumerate}[(i)]
    \item an information seeking task on the Cambridge Restaurants domain;
    \item information seeking and booking tasks for the MultiWoz domains as in ConvLab2~\cite{zhu2020convlab};
%
\item information seeking about movies within a WikiData extract;
\item a complex HelpDesk task for the DATCHA-TV~\cite{nasr2016syntactic} domain.
\end{enumerate}

\section{Conclusion and Future Work}
\label{s:conclusion}
We presented a dialogue framework in which the semantic representation embeds RDFs enabling more expressivity and scalability. The framework contains three distinct KBs: the dialogue, the task and the domain, allowing to plug in and out task and domain KBs. Moreover, we implemented four applications within the framework from single task and domain, multiple tasks and domains to complex tasks and WEB domains.

An important research question remains open: how to  integrate hierarchical representations in DST and DP? A richer representation will introduce dependence statistics and causality in these tasks. This is an important research path that might guide us towards more explainable dialogue systems.

\bibliography{sample}
\end{document}